\documentclass[runningheads]{llncs}
\usepackage{graphicx}
\usepackage{comment}
\usepackage{amsmath}
\usepackage{diagbox}
\usepackage{multirow}
\usepackage{threeparttable}
\usepackage{xcolor}
\usepackage{hyperref}

\usepackage{placeins}

\newcommand{\todo}[1]{}
\renewcommand{\todo}[1]{{\color{red} TODO: {#1}}}

\usepackage{subcaption}
\usepackage[export]{adjustbox}
\usepackage{wrapfig}
\graphicspath{{images/}}

\begin{document}
\title{Web-based Melanoma Detection}
%
% \titlerunning{Web-based Melanoma Detection}
\author{SangHyuk Kim\inst{1}\orcidID{0009-0005-1323-306X} \and
Edward Gaibor\inst{2}\orcidID{0009-0009-0400-7529} \and
Daniel Haehn\inst{3}\orcidID{0000-0001-9144-3461}}
\authorrunning{S. Kim et al.}
\institute{University of Massachusetts Boston,
\email{sanghyuk.kim001@umb.edu}\\
\and
University of Massachusetts Boston,
\email{edward.gaibor001@umb.edu}
\and
University of Massachusetts Boston,
\email{daniel.haehn@umb.edu}}
\maketitle              % typeset the header of the contribution
\begin{abstract}

Melanoma is the most aggressive form of skin cancer, and early detection can significantly increase survival rates and prevent cancer spread. However, developing reliable automated detection techniques is difficult due to the lack of standardized datasets and evaluation methods. This study introduces a unified melanoma classification approach that supports 54 combinations of 11 datasets and 24 state-of-the-art deep learning architectures. It enables a fair comparison of 1,296 experiments and results in a lightweight model deployable to the web-based MeshNet architecture named Mela-D. This approach can run up to 33x faster by reducing parameters 24x to yield an analogous 88.8\% accuracy comparable with ResNet50 on previously unseen images. This allows efficient and accurate melanoma detection in real-world settings that can run on consumer-level hardware.

% --Done--
% \todo{DH: we need to refer the F1 score or some other performance metric here as well to show how good classification is }

% Melanoma is the most aggressive form of skin cancer, and early detection has great benefits in increasing survival rates and preventing cancer spread. Recent research has taken advantage of modern deep-learning architectures to detect melanomas in medical images, and several open melanoma datasets are available. However, studies show varying results tested in different environments, making it hard to see the effectiveness of algorithms, network structures, and datasets for melanoma classification. This study introduces a unified melanoma classifying approach that supports multiple combinations of datasets and networks. The framework can preprocess and test melanoma data uniformly and generate a lightweight model deployable to the web. This approach enables a fair comparison of classifiers, and our model yields increased performance.

\keywords{Skin cancer classification  \and Medical imaging application  \and Supervised learning.}
\end{abstract}
\section{Introduction}

\begin{figure}[h!]
\centering
\includegraphics[width=0.7\textwidth]{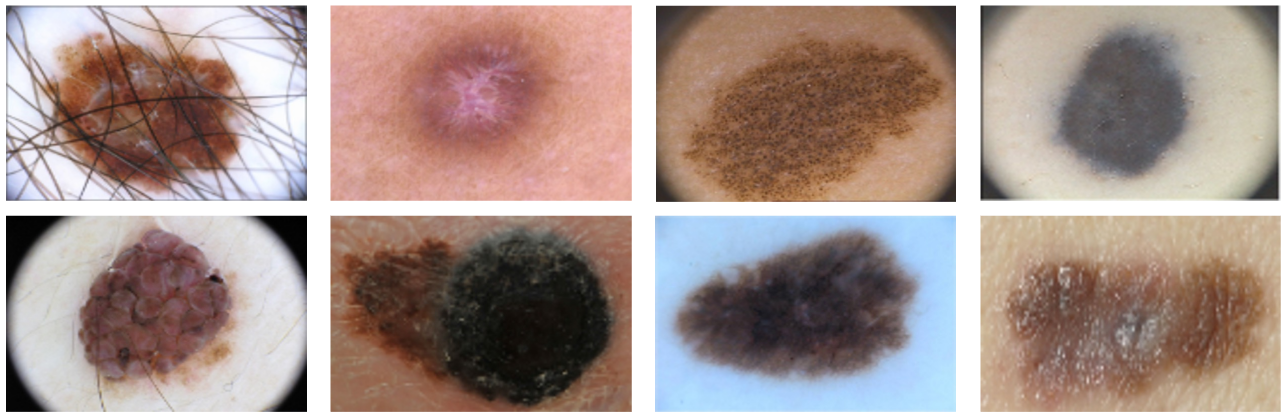}
\caption{Examples of melanoma images from different datasets illustrate the diversity in appearance and challenges for automated detection. Variations in size, shape, color, and texture across datasets highlight the need for a robust and generalizable melanoma detection approach.}
\label{fig:fig_1}
\end{figure}

\begin{figure}[h!]
\centering
\includegraphics[width=1\textwidth]{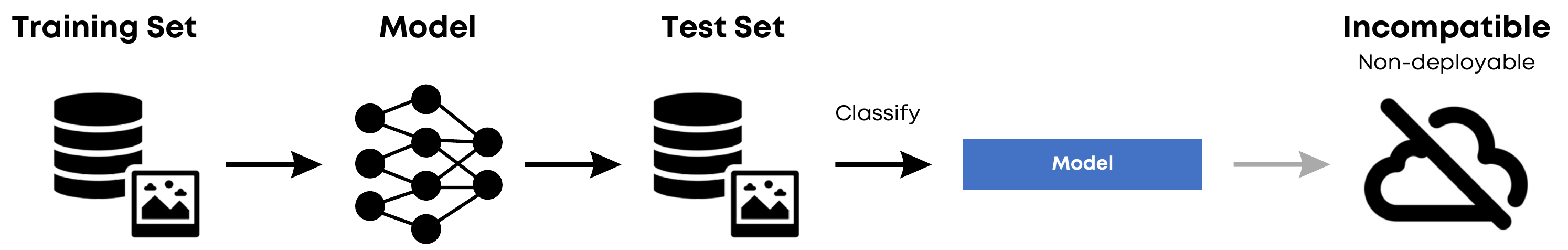}

\caption{The typical workflow of existing melanoma classifiers is non-deployable for web-based applications as each model is generated from incompatible frameworks or undergoes different preprocessing. This limits practical utility for end-users such as dermatologists.
}
\label{fig:fig_2}
\end{figure}

\begin{figure}[h!]
\centering

\includegraphics[width=1\textwidth]{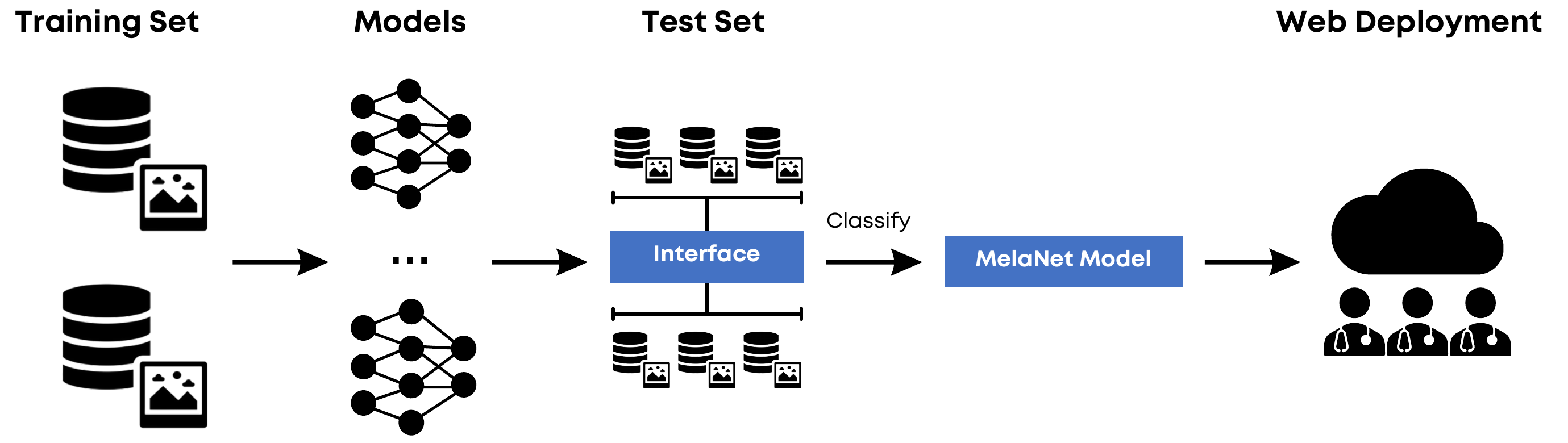}

\caption{The figure is an overview of our proposed melanoma classification framework, which is deployable on the web through a user-friendly interface. It enables end-users to easily classify melanoma images.}
\label{fig:fig_3}
\end{figure}

Melanoma is a deadly skin cancer, but early detection and treatment can increase survival rates. Convolutional neural networks (CNNs) have improved melanoma detection rates by utilizing open melanoma datasets, surpassing feature-based classifiers~\cite{starreport22}. However, developing robust and practically deployable CNN-based melanoma classifiers has several challenges. Firstly, a classifier must use diverse and representative training datasets to deal with the wide variability in melanoma appearance (Fig. ~\ref{fig:fig_1}). However, existing methods~\cite{limit1,limit2,limit3,svm,WebMelanoma} often use limited training data with non-standardized preprocessing protocols, hindering the generalizable and reproducible classifiers.
Secondly, assessing models is often inconsistent across melanoma studies, with different datasets, preprocessing methods, and performance metrics being used.
This inconsistency makes it challenging to compare each approach, as shown in (Fig. \ref{fig:fig_2}). Lastly, it is important to note that when classification models are generated from incompatible frameworks or undergo different preprocessing, they perform poorly or even become difficult to deploy (Fig. \ref{fig:fig_2}). This makes the models unusable for web-based applications, which can limit dermatologists' access to these models in real-world clinical settings.
This paper proposes a comprehensive framework for melanoma classification that enables robust evaluation and efficient web deployment (Fig. \ref{fig:fig_3}).

Our key contributions are as follows:
We curate a diverse set of 11 publicly available melanoma datasets to train and evaluate our models. This allows us to train the classifiers with various data distributions and image characteristics. We then conduct a large-scale study, training 24 state-of-the-art CNN architectures on each of the 54 dataset combinations, resulting in a total of 1,296 experiments.

% \FloatBarrier

% Figure 2

This enables us to identify the best-performing models and to assess the impact of dataset composition on classification performance. We then introduce Mela-D, a lightweight CNN architecture optimized for web deployment. Based on a MeshNet~\cite{meshnet} model, Mela-D utilizes dilated convolutions~\cite{dilation} to capture multi-scale features. Mela-D achieves a parameter reduction of up to 24x compared to conventional CNN architectures, making it suitable for real-time inference in web-based applications. We integrate Mela-D into an open-source web-based platform that allows users to upload dermatoscopic images and quickly receive real-time melanoma classification results.

\FloatBarrier
% Figure 3

\FloatBarrier

\section{Methods}
\subsection{Data Interface}

\begin{figure}[h]
\centering
\includegraphics[width=1\textwidth]{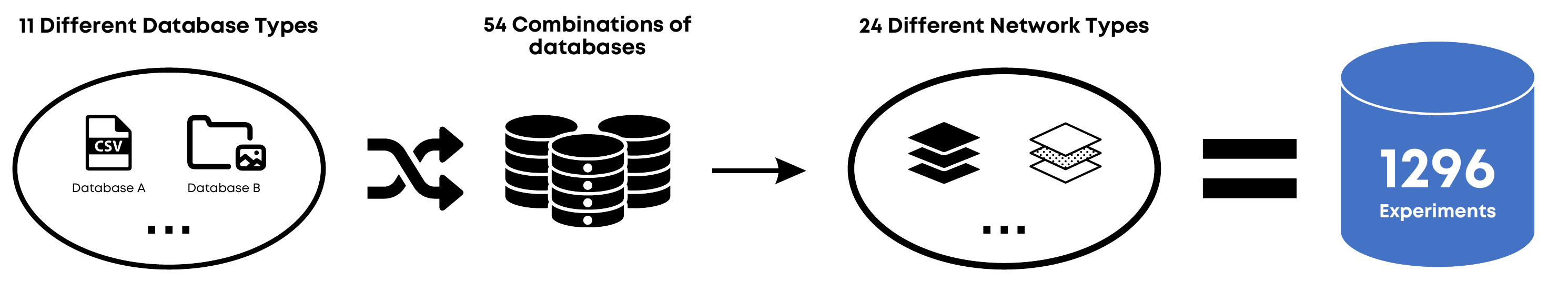}
\caption{The 11 melanoma databases containing images and labels are combined into 54 different sets. These sets are trained and evaluated with 24 network architectures, yielding 1,296 experiments.
}
\label{fig:comb_fig}
\end{figure}

The melanoma datasets~\cite{ham10000,isic2016,isic2017,isic2018,isic2019,isic2020,ph2,7pointcriteria,padufes20,mednode,kaggle} consist of JPEG and PNG images and metadata containing binary labels (benign/malignant). The image size varies between 147 x 147 and 3096 x 3096, and the directory structure varies depending on the collecting institutions. While most datasets provide ground truth in the form of comma-separated values (CSV), some datasets have folders divided for benign/malignant cases serving as labels. Our framework is capable of converting a wide range of melanoma datasets~\cite{ham10000,isic2016,isic2017,isic2018,isic2019,isic2020,ph2,7pointcriteria,padufes20,mednode,kaggle} and freely assembling them in any combination to be preprocessed and trained in CNNs. (Fig. \ref{fig:comb_fig}).

% -- Done --
% \todo{DH: the list of datasets needs to come earlier}

% \subsection{Unified Preprocessing}
% Our framework’s preprocessor unifies the skin cancer image data stored differently and performs preprocessing steps. These steps include resizing, normalization, and data augmentation. We match the ratio of images between the benign/malignant classes to be 50 : 50 using rotation, zoom in/out, random crop, and vertical/horizontal flip technique all through the open source tool Albumentation [?]. The loaded images are also normalized to prevent gradient exploding and foster effective feature learning within a network. This way, we mitigate performance decreases caused by data class imbalance and biased training.

% \clearpage
\subsection{Classification Methods}

We explore various CNN architectures as base networks to identify the best-performing melanoma classifier. The list of all CNNs tested are
\cite{densenet}: DenseNet121 / DenseNet169 / DenseNet201, \cite{efficientnet}: EfficientNet B0 to B7 \cite{inceptionv3}: InceptionV3, \cite{mobilenet}: MobileNet, \cite{mobilenetv2}:
MobileNetV2, \cite{resnet}: ResNet50 / ResNet101 / ResNet152, \cite{resnetv2}: ResNet50V2 / ResNet101V2 / ResNet152V2, \cite{vgg}: VGG16/VGG19, \cite{xception}: Xception.
We use the Adam optimizer~\cite{kingma2014adam} with 0.0001 learning rate, 32 batch size, 20 epochs, and categorical cross-entropy loss training from ImageNet pre-trained weights for all models except Mela-D. Each network's binary softmax activation function predicts the benign/malignant class probabilities. We resize every image to 150x150 pixels and apply data augmentation (rotation, zoom, cropping, flipping) to balance the benign/malignant ratio to 50:50.

\subsection{Mela-D and Web Deployment}

\begin{figure}[h]
\centering
\includegraphics[width=1\textwidth]{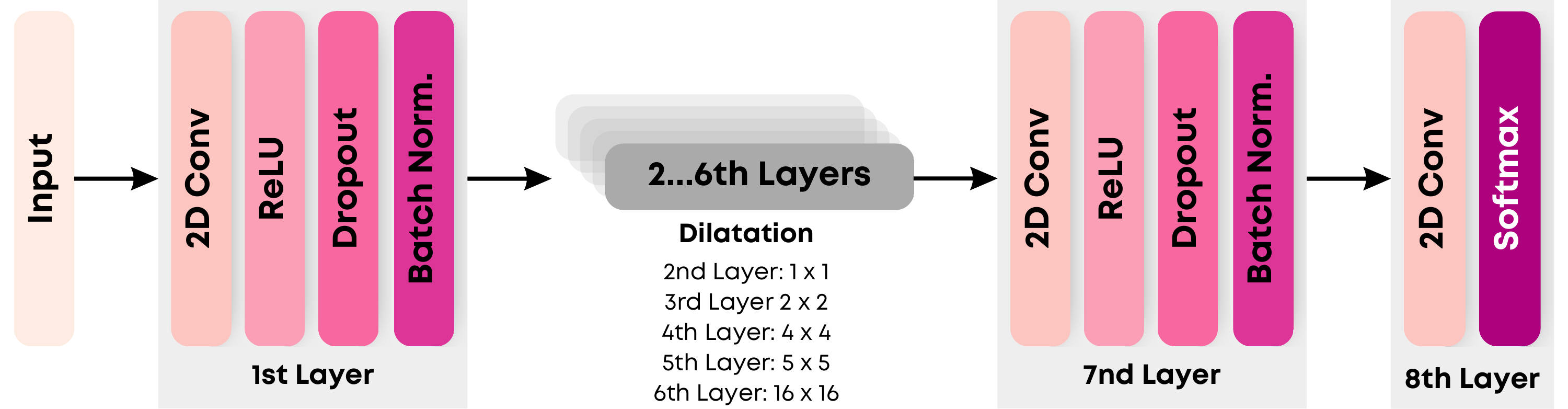}
\caption{Mela-D architecture uses Dilated convolutions~\cite{dilation} with increasing dilation factors in each layer to achieve independent computation during inference with lower computational cost.}
\label{fig:fig_4}
\end{figure}

MeshNet architecture~\cite{meshnet} employs a compact CNN model using dilation convolution~\cite{dilation} to flexibly control the size of the receptive field of the input neuron. In this regard, the model can integrate multiple contexts without increasing the number of layers and parameters while maintaining comparable performance to the existing CNNs. Dilated convolution is represented as:

% --Done--
% \todo{DH: cite https://www.vis.xyz/pub/dilation/}

\begin{equation}
\left(F*_lk\right)\left(\mathbf{p}\right)=\sum_{\mathbf{s}+l\mathbf{t}=p} F\left(\mathbf{s}\right)k\left(\mathbf{t}\right)
\end{equation}

where $l$ is dilation factor and $*_l$ is $l$-dilated convolution. Based on MeshNet~\cite{meshnet}, we build a Mela-D architecture (Fig.~\ref{fig:fig_4}) to reduce the number of parameters by 24x compared to the existing network by flexibly controlling the receptive field size. All the layers independently perform inference and do not lose information in feed-forward.

% -- Done --
% \todo{DH: cite https://www.vis.xyz/pub/dilation/}

% \clearpage

\section{Experiments and Results}

We conduct experiments using Python 3.9, Keras 2.5.0rc0, TensorFlow 2.5.0, and high-performance computing resources (AMD EPYC 7742 CPUs, Nvidia DGX-A100 GPUs, 2064GB RAM). Runtime benchmarks are performed using Google Chrome Performance Monitor on a Windows 11 laptop (Intel Core i5-1135G7 @ 2.40GHz, Intel Iris Xe Graphics, 8GB RAM) with WebGL TensorFlow.js backend.

\subsection{Performance}

We comprehensively evaluate 1,296 experiments on five test sets~\cite{ham10000,isic2016,isic2017,isic2018,7pointcriteria} using various performance metrics such as precision, sensitivity, specificity, F1-score, and accuracy. The models are ranked based on precision, the primary metric used in the ISIC challenge~\cite{isic2016}. We then select the top three models for each test set based on this ranking metric. As can be seen in Table \ref{table:1},
ResNet152~\cite{resnet}, ResNet50~\cite{resnet}, and DenseNet121~\cite{densenet} are consistently among the best-performing classifiers.

% \FloatBarrier

\begin{table}[h!]{
% \caption{Precision vs. runtime time ratio comparison for MelaNet and the top-performing models from each test set. The ratio is calculated as the model's precision divided by its average scripting time in seconds. MelaNet achieves the highest ratio, indicating an optimal balance between classification accuracy and execution speed. This makes MelaNet well-suited for efficient web-based deployment and real-time inference}
\caption{Top-3 performing models per combination of dataset and classifier. The performance is calculated based on precision, specificity, sensitivity, and accuracy. The results are in precision descending order. The best-performing metric per test set is in \textbf{bold}.}
\centering

\begin{adjustbox}{width=1\textwidth}{
\begin{threeparttable}{
\footnotesize
\begin{tabular}{c p{2.0cm} p{2cm} p{1.5cm} p{1.5cm} p{1.5cm} p{1.5cm} }
%\begin{tabular}{@{} p{2cm}|p{4cm}|p{3cm}|p{3cm} @{}}

\hline
\multicolumn{1}{c}{Testset} & \multicolumn{1}{c}{Classifier} & \multicolumn{1}{c}{Trainsets} & \multicolumn{1}{c}{Precision} & \multicolumn{1}{c}{Specificity} & \multicolumn{1}{c}{Sensitivity} & \multicolumn{1}{c}{Accuracy}  \\ \hline

\multirow{3}{*}{HAM10000~\cite{ham10000}}  & \multicolumn{1}{c}{ResNet152 \cite{resnet}} & \multicolumn{1}{c}{\tiny a+b+c+d+e+g} & \multicolumn{1}{c}{\textbf{99.3\%}} &
\multicolumn{1}{c}{\textbf{100\%}} &
\multicolumn{1}{c}{\textbf{70.6\%}} & \multicolumn{1}{c}{\textbf{98.6\%}}   \\

& \multicolumn{1}{c}{VGG16 \cite{vgg}}       & \multicolumn{1}{c}{\tiny a+b+c+d+e+f+g}       & \multicolumn{1}{c}{99.1\%}       & 
\multicolumn{1}{c}{\textbf{100\%}}       & 
\multicolumn{1}{c}{62.7\%}          & \multicolumn{1}{c}{98.3\%}   \\

& \multicolumn{1}{c}{EfficientNetB2 \cite{efficientnet}}       & \multicolumn{1}{c}{\tiny a+b+c+d+e}       & \multicolumn{1}{c}{99.0\%} &  \multicolumn{1}{c}{\textbf{100\%}}      
& \multicolumn{1}{c}{60.8\%}
 
& \multicolumn{1}{c}{98.2\%}    \\ \cline{1-7}

% & \multicolumn{1}{c}{VGG16}       & \multicolumn{1}{c}{\tiny1+2+3+4+5+6+7+8}       & \multicolumn{1}{c}{99.0\%}       & \multicolumn{1}{c}{58.8\%}         & \multicolumn{1}{c}{98.0\%}    \\ \cline{1-6}

\multirow{3}{*}{ISIC2016~\cite{isic2016}}  & \multicolumn{1}{c}{ResNet152~\cite{resnet}} & \multicolumn{1}{c}{\tiny a+b+c+d+e+g} & \multicolumn{1}{c}{\textbf{96.3\%}} &
\multicolumn{1}{c}{\textbf{100\%}} &
\multicolumn{1}{c}{68.0\%} & \multicolumn{1}{c}{93.7\%}   \\

& \multicolumn{1}{c}{ResNet152 \cite{resnet}}       & \multicolumn{1}{c}{\tiny a+b+c+d+e}       & \multicolumn{1}{c}{96.2\%}       & 
\multicolumn{1}{c}{99.7\%}   & 
\multicolumn{1}{c}{\textbf{74.7\%}}   & \multicolumn{1}{c}{\textbf{94.7\%}}   \\

& \multicolumn{1}{c}{ResNet50 ~\cite{resnet}}       & \multicolumn{1}{c}{\tiny a+b+c+d+h+i}       & \multicolumn{1}{c}{96.0\%}     & 
\multicolumn{1}{c}{99.7\%}       & 
\multicolumn{1}{c}{73.3\%}       & \multicolumn{1}{c}{94.4\%}    \\ \cline{1-7}
                                  % & \multicolumn{1}{c}{ResNet101}       & \multicolumn{1}{c} \cline{1-6}{\tiny1+2+3+4+5+6+7+9+10}       & \multicolumn{1}{c}{96.0\%}       & \multicolumn{1}{c}{73.3\%}         & \multicolumn{1}{c}{94.5\%}    \\ \cline{1-6}
\multirow{3}{*}{ISIC2017~\cite{isic2017}}  & \multicolumn{1}{c}{ResNet152\cite{resnet}} & \multicolumn{1}{c}{\tiny d} & \multicolumn{1}{c}{\textbf{91.3\%}} & \multicolumn{1}{c}{99.2\%} & \multicolumn{1}{c}{49.6\%} & \multicolumn{1}{c}{\textbf{89.5\%}}   \\

& \multicolumn{1}{c}{VGG19\cite{vgg}}       & \multicolumn{1}{c}{\tiny a+b+c+d+e+h+i}       & \multicolumn{1}{c}{91.3\%}       & 
\multicolumn{1}{c}{98.8\%}       & 
\multicolumn{1}{c}{56.4\%}          & \multicolumn{1}{c}{90.5\%}   \\

& \multicolumn{1}{c}{EfficientNetB0 \cite{vgg}}       & \multicolumn{1}{c}{\tiny a+f}       & \multicolumn{1}{c}{90.4\%}       & 
\multicolumn{1}{c}{\textbf{100\%}}         & 
\multicolumn{1}{c}{2.0\%}         & \multicolumn{1}{c}{80.8\%}    \\ \cline{1-7}
                                  % & \multicolumn{1}{c}{VGG19}       & \multicolumn{1}{c}{\tiny1+2+3+4+5+7+8+9}       & \multicolumn{1}{c}{90.1\%}       & \multicolumn{1}{c}{\textbf{61.6\%}}         & \multicolumn{1}{c}{91.0\%}    \\ \cline{1-6}
\multirow{3}{*}{ISIC2018~\cite{isic2018}}  & \multicolumn{1}{c}{ResNet50\cite{resnet}} & \multicolumn{1}{c}{\tiny c} & \multicolumn{1}{c}{\textbf{76.1\%}}& \multicolumn{1}{c}{\textbf{97.3\%}} & \multicolumn{1}{c}{32.2\%} & \multicolumn{1}{c}{89.9\%}   \\

& \multicolumn{1}{c}{DenseNet201\cite{densenet}}       & \multicolumn{1}{c}{\tiny a+b+c+d+e+f+h}       & \multicolumn{1}{c}{75.9\%}       & 
\multicolumn{1}{c}{96.2\%}          & 
\multicolumn{1}{c}{42.7\%}          & \multicolumn{1}{c}{\textbf{90.1\%}}   \\
                                  
& \multicolumn{1}{c}{DenseNet121\cite{densenet}}       & \multicolumn{1}{c}{\tiny e}       & \multicolumn{1}{c}{75.9\%}       & 
\multicolumn{1}{c}{99.6\%}         & 
\multicolumn{1}{c}{5.8\%}         & \multicolumn{1}{c}{89.0\%}    \\ 
\cline{1-7}                                 
                                  % & \multicolumn{1}{c}{ResNet152}       & \multicolumn{1}{c}{\tiny1+3+4+5}       & \multicolumn{1}{c}{75.9\%}       & \multicolumn{1}{c}{\textbf{35.7\%}}         & \multicolumn{1}{c}{89.9\%}    \\ \cline{1-6}
\multirow{3}{*}{7-point criteria~\cite{7pointcriteria}}  & \multicolumn{1}{c}{ResNet50\cite{resnet}} & \multicolumn{1}{c}{\tiny e} & \multicolumn{1}{c}{\textbf{87.3\%}}  & \multicolumn{1}{c}{\textbf{100\%}} & \multicolumn{1}{c}{0.9\%} & \multicolumn{1}{c}{74.7\%}   \\

& \multicolumn{1}{c}{ResNet50\cite{resnet}}       & \multicolumn{1}{c}{\tiny a+b+c+h+g}       & \multicolumn{1}{c}{33.7\%}       & 
\multicolumn{1}{c}{98.0\%}   & 
\multicolumn{1}{c}{\textbf{81.5\%}}   & \multicolumn{1}{c}{74.7\%}   \\
                                  
& \multicolumn{1}{c}{DenseNet121\cite{densenet}}       & \multicolumn{1}{c}{\tiny a+f+h}       & \multicolumn{1}{c}{82.2\%}       & 
\multicolumn{1}{c}{99.0\%}         & 
\multicolumn{1}{c}{18.8\%}         & \multicolumn{1}{c}{\textbf{78.5\%}}    \\ \cline{1-7}
                                  
                                  % & \multicolumn{1}{c}{EfficientNetB2}       & \multicolumn{1}{c}{\tiny1+5+7}       & \multicolumn{1}{c}{82.2\%}       & \multicolumn{1}{c}{18.8\%}         & \multicolumn{1}{c}{\textbf{78.5\%}}    \\ \cline{1-6}
\end{tabular}
\caption*{\scriptsize a:ISIC16~\cite{isic2016} b:ISIC2017~\cite{isic2017} c:ISIC2018~\cite{isic2018} d:ISIC2019~\cite{isic2017} e:ISIC2020~\cite{isic2020} \\f:7-point criteria~\cite{7pointcriteria} g:PH2~\cite{ph2} 
h:PAD\_UFES\_20\cite{padufes20} i: MEDNODE~\cite{mednode} \\j: Kaggle~\cite{kaggle}}

}\end{threeparttable}
}
\end{adjustbox}

\label{table:1}
}\end{table}

% --Done--
% \todo{DH: what about specificity - why nt report that?}

However, we can observe that the best-performing models have slower runtimes on the browser than the web-deployable Mela-D (Table \ref{table:2}).

\begin{table}[h!]
\caption{We measure the runtime of the best-performing on the browser. Each classifier has the highest precision on each test set. We conduct three trials to measure the runtimes. The fastest runtime is in \textbf{bold}. Our observations show that our proposed model, Mela-D, is the fastest classifier among all the ones we test. }

% --Done--
% \todo{DH: here the footnote needs to include the citations again like table 1}

\centering

\begin{adjustbox}{width=1\textwidth}
\begin{threeparttable}
\footnotesize
\begin{tabular}{c p{2.3cm} p{2cm} p{1.5cm} p{1.5cm} p{1.5cm} p{1cm} }
%\begin{tabular}{@{} p{2cm}|p{4cm}|p{3cm}|p{3cm} @{}}

\hline
\multicolumn{1}{c}{Classifier} & \multicolumn{1}{c}{Trainsets} & \multicolumn{1}{c}{\begin{tabular}[c]{@{}c@{}} 1st trial\end{tabular}} & \multicolumn{1}{c}{\begin{tabular}[c]{@{}c@{}} 2nd trial\end{tabular}} & \multicolumn{1}{c}{\begin{tabular}[c]{@{}c@{}} 3rd trial\end{tabular}} & \multicolumn{1}{c}{\begin{tabular}[c]{@{}c@{}} Average\end{tabular}} \\  \hline
\multirow{1}{*}{Proposed method}  & \multicolumn{1}{c}{\tiny a+b+c+d+e+g+f+i+j+k} & \multicolumn{1}{c}{\textbf{682 ms}} & \multicolumn{1}{c}{\textbf{654 ms}} & \multicolumn{1}{c}{\textbf{621 ms}} & \multicolumn{1}{c}{\textbf{652.3 $\pm$ 30.5 ms}}    \\

\multirow{1}{*}{ResNet152~\cite{resnet}}  & \multicolumn{1}{c}{\tiny a+b+c+d+e+g} & \multicolumn{1}{c}{9125 ms} & \multicolumn{1}{c}{6129 ms} & \multicolumn{1}{c}{7177 ms} & \multicolumn{1}{c}{7477 $\pm$ 1520.4 ms}   \\

\multirow{1}{*}{DenseNet121~\cite{densenet}}  & \multicolumn{1}{c}{\tiny a+b+c+d+i+j} & \multicolumn{1}{c}{22204 ms} & \multicolumn{1}{c}{21987 ms} & \multicolumn{1}{c}{20683 ms } & \multicolumn{1}{c}{21624.7 $\pm$ 822.7 ms}   \\

\multirow{1}{*}{ResNet50~\cite{resnet}}  & \multicolumn{1}{c}{\tiny c} & \multicolumn{1}{c}{1791 ms} & \multicolumn{1}{c}{1438 ms} & \multicolumn{1}{c}{1663 ms} & \multicolumn{1}{c}{1630.7 $\pm$ 178.7 ms}   \\

\multirow{1}{*}{ResNet152~\cite{resnet}}  & \multicolumn{1}{c}{\tiny d} & \multicolumn{1}{c}{12065 ms} & \multicolumn{1}{c}{9328 ms} & \multicolumn{1}{c}{12017 ms} & \multicolumn{1}{c}{11136.7 $\pm$ 1566.5 ms}   \\

\multirow{1}{*}{ResNet50~\cite{resnet}}  & \multicolumn{1}{c}{\tiny e} & \multicolumn{1}{c}{7264 ms} & \multicolumn{1}{c}{7546 ms} & \multicolumn{1}{c}{7669 ms} & \multicolumn{1}{c}{7493 $\pm$ 207.6 ms}   \\ \hline

\end{tabular}
\caption*{\scriptsize a:ISIC16~\cite{isic2016} b:ISIC2017~\cite{isic2017} c:ISIC2018~\cite{isic2018} d:ISIC2019~\cite{isic2019} e:ISIC2020~\cite{isic2020} f:7-point criteria~\cite{7pointcriteria} g:PH2~\cite{ph2} \\
h:PAD\_UFES\_20~\cite{padufes20} i:MEDNODE~\cite{mednode} j:Kaggle~\cite{kaggle}}

\end{threeparttable}
\end{adjustbox}
\label{table:2}
\end{table}

\begin{figure}[h!]
\centering
\includegraphics[width=0.8\textwidth]{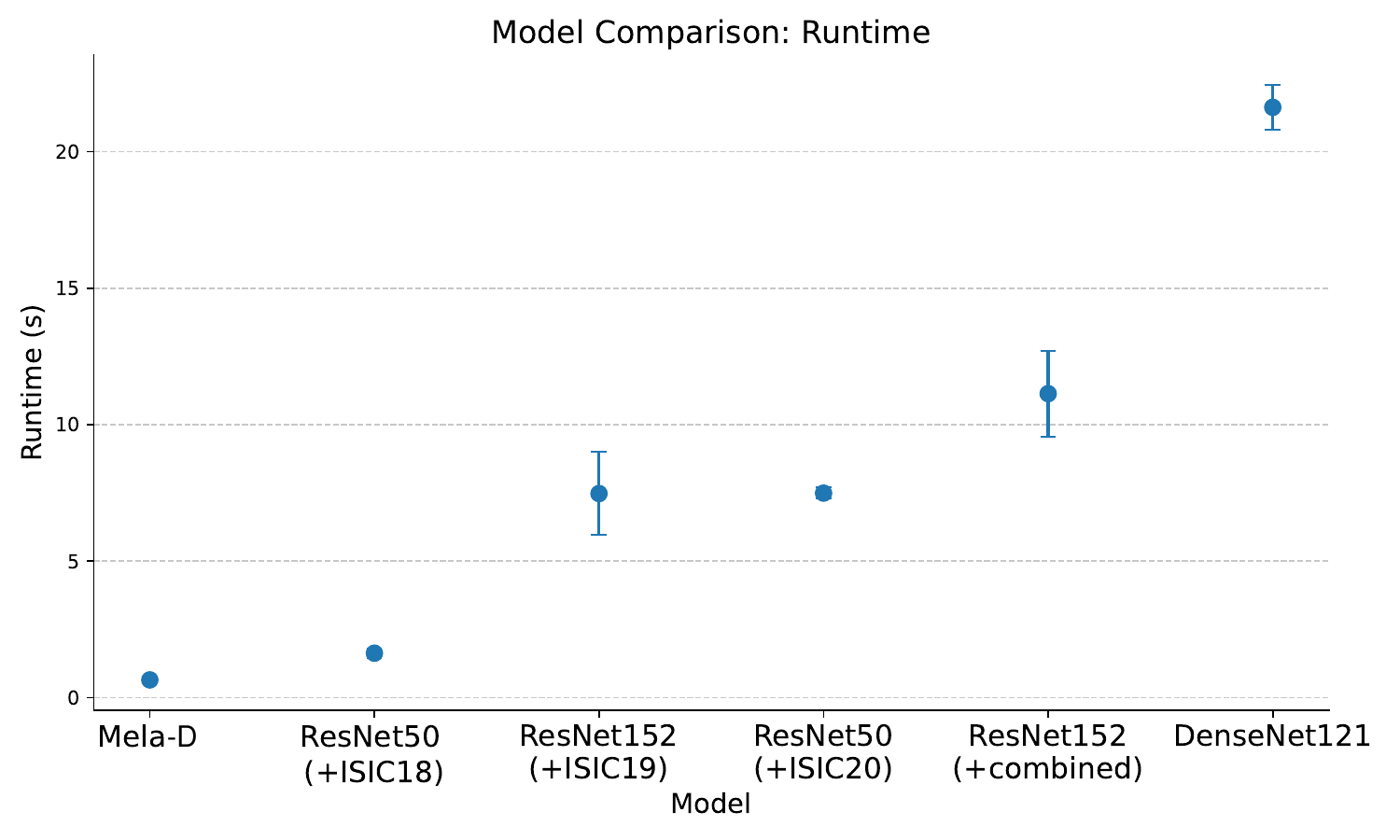}
% execution time OR runtime
\caption{Comparison of average execution time (in seconds) for Mela-D and the top-performing models from each test set. Mela-D demonstrates 2 to 33 times faster runtimes than the other models. Tested with Google Chrome on Intel Core i5-1135G7 @ 2.40GHz, Intel Iris Xe Graphics, 8GB RAM, with WebGL TensorFlow.js backend.}
\label{fig:fig_6}
\end{figure}

\FloatBarrier
To investigate this aspect, we deploy the models on the web and perform three execution trials, calculating the average runtime in seconds. The results in (Fig. \ref{fig:fig_6}) show that Mela-D outperforms the other models, demonstrating a minimum of twice and a maximum of 33 times faster execution on the browser.

\FloatBarrier

% \begin{figure}[h!]
% \centering
% % \includegraphics[width=11cm,height=6.2cm]{images/Tab3Plot_PerformanceRatio.pdf}
% \includegraphics[width=0.5\textwidth]{images/Tab3Plot_PerformanceRatio.pdf}
% \caption{The selected models are the best-performing models in terms of precision on 5 test sets, and MelaNet is presented to compare. \todo{DH: weird warped graphic, only specify height or width} Scripting time is obtained from the web browser, and then the ratio is calculated by dividing model performance by the amount of time for scripting. MelaNet is the most cost-effective when running on the web}
% \label{fig:fig_6}
% \end{figure}

\FloatBarrier
\begin{figure}[t!]
\centering
\includegraphics[width=0.6\textwidth]{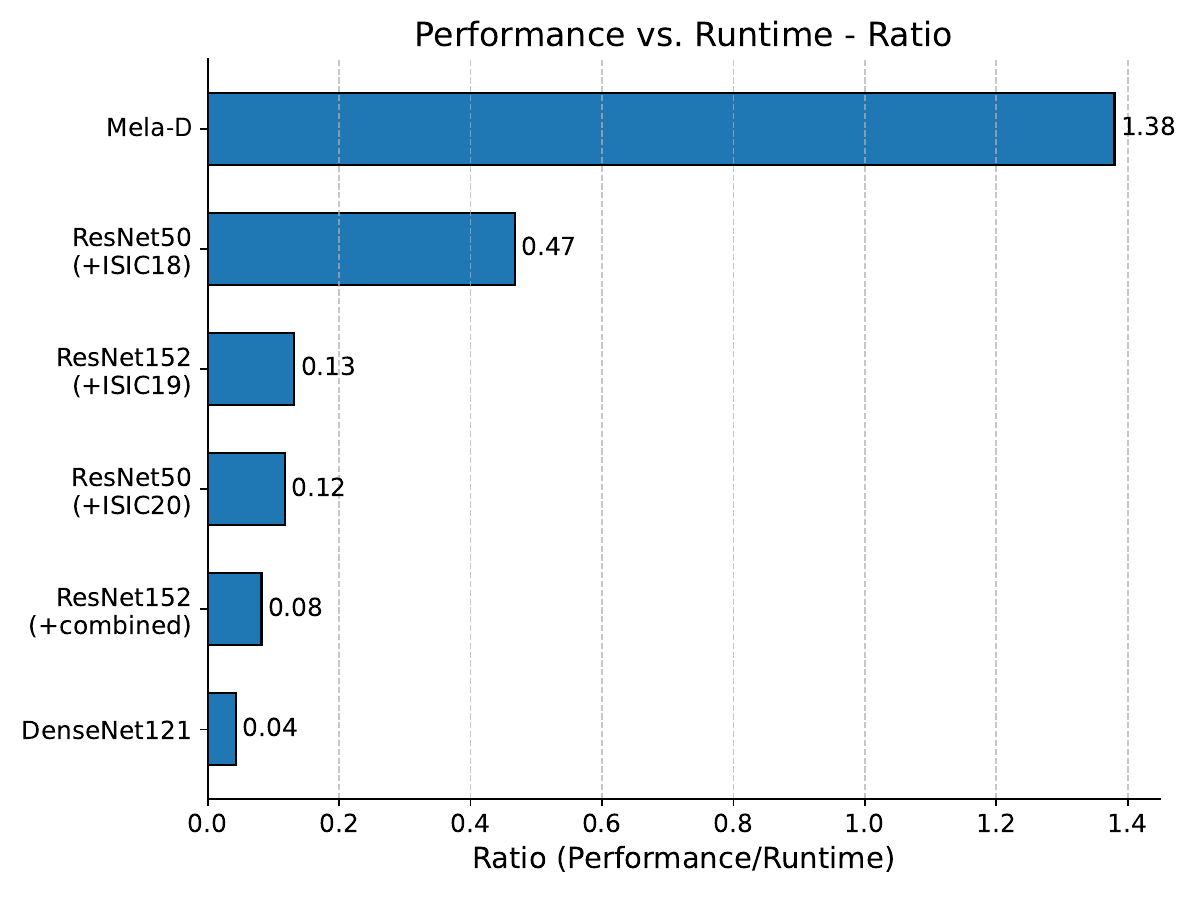}
\caption{ Mela-D achieves the highest ratio, indicating a high balance between classification accuracy and runtime speed. Thus, Mela-D is well-suited for efficient web-based deployment and real-time inference.}
\label{fig:fig_7}
\end{figure}

To evaluate the scalability of Mela-D when considering running it on consumer devices, we additionally analyze the ratio between performance and runtime. We assess the performance (precision) vs. runtime ratio of Mela-D and the top-performing models from each test set to verify the efficiency of performance versus runtime. The formula divides precision by its average runtime in seconds. The results are shown in (Table \ref{table:2}). The analysis reveals that Mela-D achieves the highest performance-time ratio among the best-performing models with accuracy 88.8\% on ISIC18~\cite{isic2018} benchmark and 652 ms runtime on the web. At the same time, ResNet50~\cite{resnet} records accuracy 89.9\% on ISIC18\cite{isic2018} and 1630 ms runtime on the web, indicating Mela-D's superiority in performance versus computational efficiency (Fig. \ref{fig:fig_7}).

\FloatBarrier
\begin{figure}[h!]
\centering
\includegraphics[width=1\textwidth]{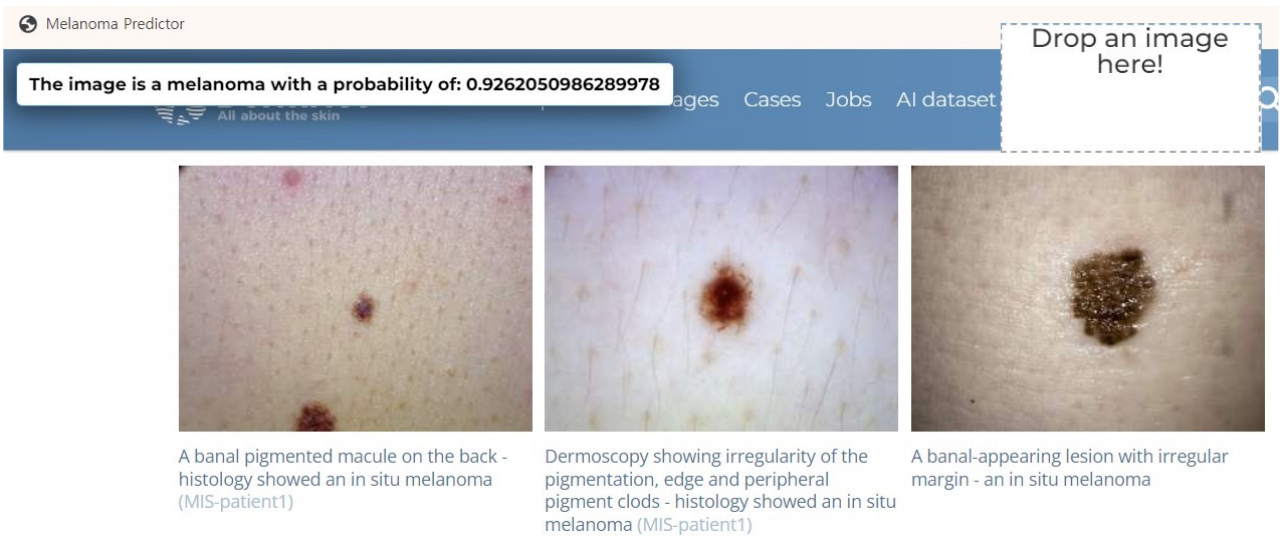}
\caption{ Mela-D, deployed on Boostlet.js, a web-based image processing plugin platform, performs melanoma prediction on images on the web with a drag-and-drop interface. }
\label{fig:boostlet}
\end{figure}

\subsection{Web Deployment}

\begin{figure}[h!]
\centering
\includegraphics[width=0.57\textwidth]{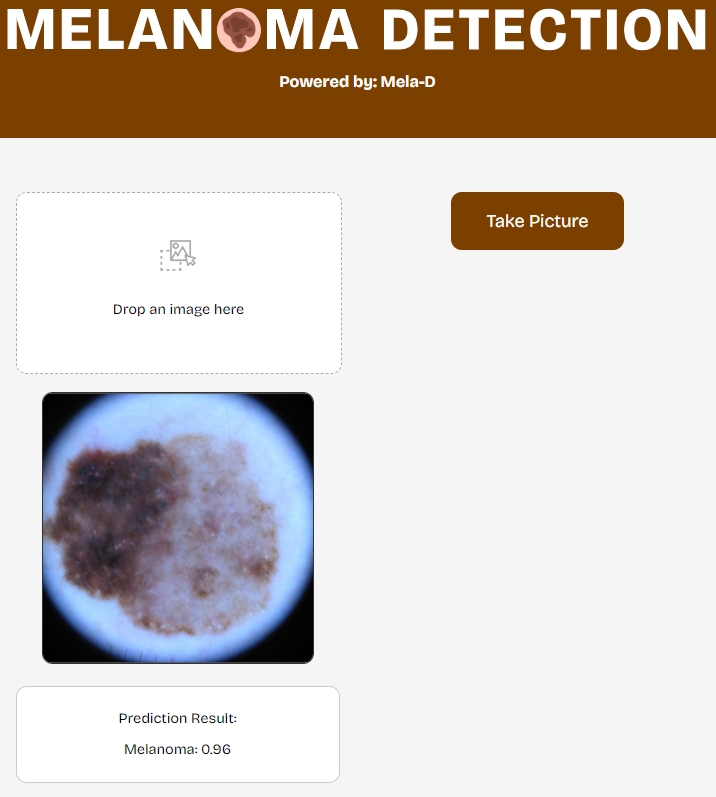}
\caption{ Users can determine skin cancer from any image they capture through a standalone web application.  }
\label{fig:standalone_web}
\end{figure}

% Web Deployment
We deploy Mela-D to the web through integration with the open-source platform at \href{https://boostlet.org}{https://boostlet.org},
allowing client-side image processing via custom browser bookmarklets that inject JavaScript functionality when visiting websites.
% \todo{DH: add examples such as the NIC Imaging data commons, i think mentioned that before, cite https://pubmed.ncbi.nlm.nih.gov/34185678/ and also openneuro}. <<<< not applicable anymore, i added ermnetnz.org below
\noindent
We convert Mela-D into a TensorflowJS layers model to exploit its lightweight nature through a simple drag-and-drop interface. When visiting a website containing skin images such as DermNetNZ.org~\cite{DermNet}, users can click the bookmarklets, drag an image onto the page, and receive a melanoma classification result via a front-end alert (see Fig. \ref{fig:boostlet} and supplementary video). We also have a standalone web application where users can test local images at \href{https://mpsych.github.io/melanoma/}{https://mpsych.github.io/melanoma/} as shown in Fig. \ref{fig:standalone_web}.

\section{Conclusions}
We have introduced a melanoma detection framework with a lightweight machine learning model, Mela-D, to leverage existing databases to build a robust and compact classifier for the web. We verify its efficiency by comparing the performance versus runtime ratio with existing CNNs and deploying the model on the web. In the future, we plan to build an affordable skin change tracker based on this framework. We release our work and results as open source at \href{https://mpsych.org/melanoma}{https://mpsych.org/melanoma}.

\bibliographystyle{splncs04}
\bibliography{references.bib}
% \bibitem{ref_article2}




% ISIC Melanoma Challenge, \url{https://challenge.isic-archive.com/landing/2016/39/}. Last accessed 6
% Mar 2024
% \end{thebibliography}
\end{document}